\PassOptionsToPackage{table}{xcolor}
\documentclass{fairmeta}
\usepackage{lmodern}

\microtypesetup{expansion=false}

\usepackage[utf8]{inputenc}
\usepackage{url}
\usepackage{amsfonts}
\usepackage{nicefrac}
\usepackage{floatrow}
\usepackage{xspace}
\usepackage{array}
\usepackage{soul}
\usepackage{wrapfig}
\usepackage{bbm}
\usepackage{amsmath}
\usepackage{amssymb}
\usepackage{mathtools}
\usepackage{pifont}
\usepackage{multicol}
\usepackage{listings}
\usepackage[accsupp]{axessibility}
\def\ours{\textbf{{VideoABC}}\@\xspace}

\def\oursben{\textbf{{VideoABC-Bench}}\@\xspace}
\def\judge{\textbf{{Judge}}\@\xspace}
\def\direct{\textbf{{Direct}}\@\xspace}
\def\mlp{\textbf{{MLP-2L}}\@\xspace}
\def\mlpa{\textbf{{MLP-A}}\@\xspace}

\def\and{\textit{and}\@\xspace}
\def\ood{\textbf{{Eval-OOD}}\@\xspace}
\def\id{\textbf{{Eval-ID}}\@\xspace}
\def\ie{\textit{i.e.}\@\xspace}
\def\eg{\textit{e.g.}\@\xspace}

\title{An Attribute-Based Measure of Video Complexity}
\author[2]{Aditya Sarkar}
\author[3,\dagger]{Yi Li}
\author[1]{Zihao Wang}
\author[1]{Hao Yuan}
\author[1]{Sai Vidyaranya Nuthalapati}
\author[1]{Aashu Singh}
\author[4]{Jiacheng~Cheng}
\author[2]{Ang~Li}
\author[1]{Shlok Kumar Mishra}
\author[2]{David Jacobs}
\author[3]{Nuno Vasconcelos}

\affiliation[1]{Meta AI}
\affiliation[2]{University of Maryland College Park}
\affiliation[3]{University of California San Diego}
\affiliation[4]{Yale~University}

\date{\today}
\correspondence{\email{shlokmishra@meta.com}, \email{jiacheng.cheng@yale.edu}}

\abstract{\label{sec:abstract}
A new framework for the estimation of the complexity posed by video-question pairs to video-LLMs, {\it Video Attribute-Based Complexity} (\ours), is proposed. Video complexity is defined as the probability of failure of a video-LLM for a given video-question pair. \ours is a non-parametric complexity measure, using a reference video dataset and a pre-defined vocabulary of video attributes informative of complexity, \eg the scene complexity or the speed of the video event relevant to the question. In a training phase, reference videos are projected into the space of these attributes, which is then quantized. The expected ABC of each quantization cell is then computed. Given a new video and its projection into the attribute space, complexity is estimated by the  expected ABC of the associated quantization cell. To enable the use of \ours with small reference video datasets, two quantizers are combined: a k-means quantizer that enables accurate complexity estimates for samples in the distribution of the reference dataset and a universal lattice quantizer that guarantees generalization to out-of-distribution samples. A synthetic video generation procedure, inspired by target-distractor manipulations of psychophysics studies, is proposed to populate the cells of the lattice quantizer during training, enabling the computation of their expected ABCs. Experimental results show that \ours is effective even with very low-dimensional attribute representations, substantially outperforming approaches like `video-LLM as judge' with much less complexity. Finally, the explainable nature of the \ours score, in terms of well-defined attributes, is shown to provide insights on how the attribute composition of benchmarks affects their complexity. 
}

\newtcolorbox{findingbox}{
    colback=metabg,
    colframe=metablue,
    boxrule=1pt,
    arc=6pt,
    left=8pt,
    right=8pt,
    top=6pt,
    bottom=6pt
}

\begin{document}
\maketitle

\begingroup
\renewcommand\thefootnote{\fnsymbol{footnote}}
\footnotetext[2]{Work done at UC San Diego. Currently working at Qualcomm.} 
\endgroup

\section{Introduction}
\label{sec:intro}

In recent years, there have been significant advances in Large Language Models for video (video-LLMs)~\citep{yang2025qwen3, zhu2025internvl3, chen2023videollm}, namely in areas like video-question answering~\citep{zhou2024mlvu, li2024mvbench, wu2024longvideobench}. However, these models are expensive and far from perfect. An important open question for the development of both better models and more effective applications is how to predict the probability of failure of a video-LLM for a particular video-question pair. In this work, we refer to this probability as a measure, or score, of video complexity.  
In the absence of complexity measures, even the simple task of benchmarking model performance can be extremely inefficient. For example, \citep{zohar2025apollo, feng2025breaking} report that the 184 A100 GPU hours required to evaluate a small (3B-parameter) model across all existing benchmarks are mostly spent on video-question pairs answerable by an LLM without video input, providing no insight about the video representation. This has motivated the research community to compute the complexity of video-question pairs and test models only on those that are hard \citep{fu2025video}. 

Existing approaches to estimating complexity \citep{platanios2019competence, spitkovsky2009baby} are usually parametric, relying on a model, typically an LLM, to produce a holistic complexity score. For example, \citep{graves2016adaptive} measures video complexity by counting the number of reasoning steps a video-LLM takes to answer a question. Similarly, \citep{eyzaguirre2025understanding} proposes generating executable code with an LLM to solve the query and uses this process as a proxy for difficulty. However, these approaches have several limitations. First, they do not utilize video attributes to determine its video complexity. It is well-known in the literature that attributes determine complexity~\citep{li2024mvbench, wu2024longvideobench, zhou2024mlvu} \eg a long video or a video with a fast event is complex. Yet existing complexity models do not rely on them to determine their score thereby limiting the diagnostic value of these methods. Second, they are computationally inefficient. There are many applications where it would be useful to obtain a complexity score {\it before} the video is processed by the video-LLM. In fact, a common approach to evaluate model performance, usually referred to as {\it LLM-as-judge} (denoted \judge in this work) is to rely on an external model to evaluate the performance of the target model~\citep{eyzaguirre2025understanding, platanios2019competence}. While these approaches can produce complexity scores, effective performance usually requires \judge models much larger and more complex than the target model, typically proprietary models like GPT~\citep{achiam2023gpt} or Gemini~\citep{pichai2025new}. Such a model may be infeasible in applications where computational resources are limited. 
Thus there is a pressing need for complexity scores that are interpretable, predictive and computationally efficient.


\begin{figure*}[t]\RawFloats
    \centering
    \begin{tabular}{@{}c@{}}
        \includegraphics[width=\textwidth]{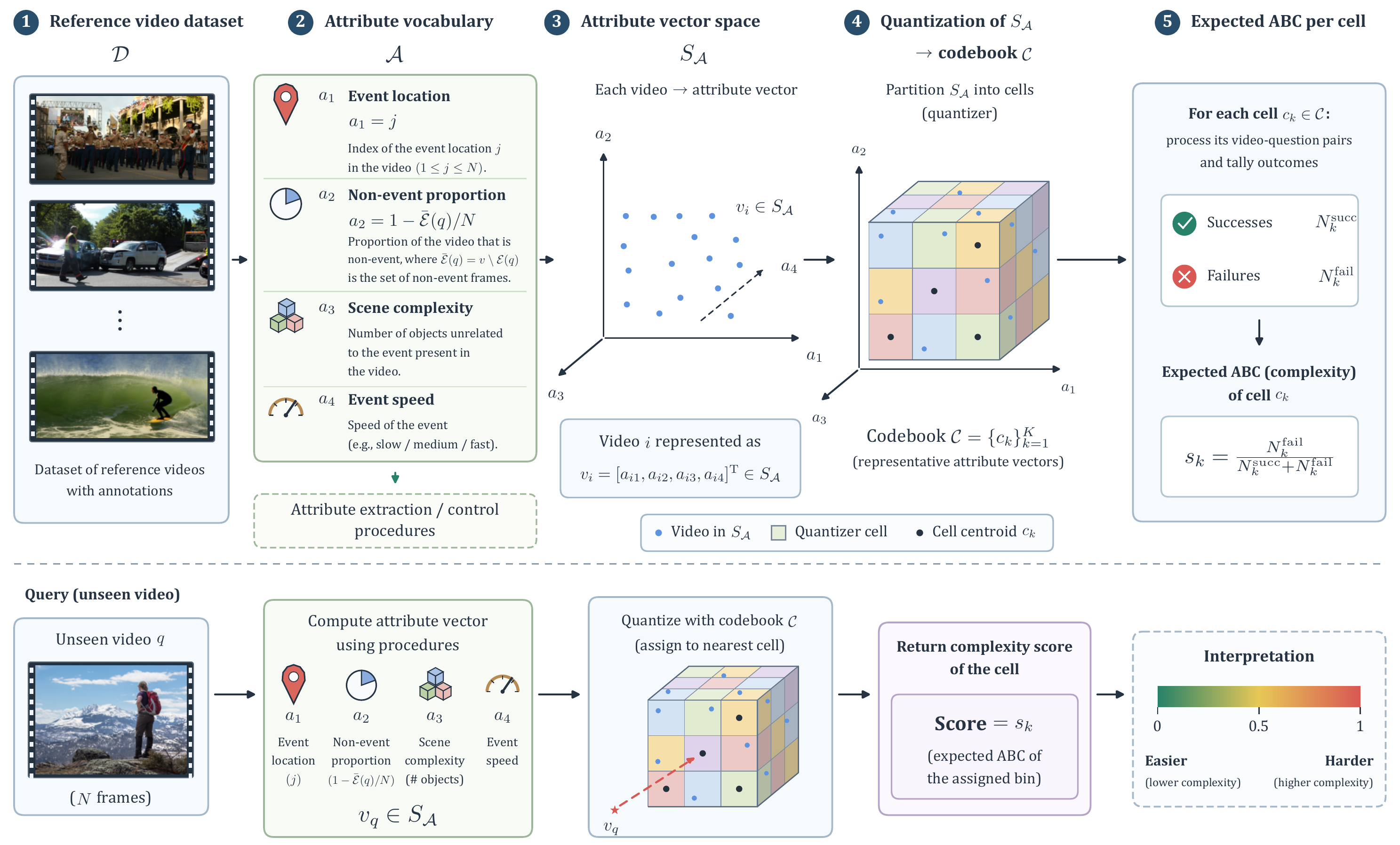}
    \end{tabular}
    \caption{\ours is an approach for the  characterization of video-question complexity in terms of a set of video attributes. Top: 1) given a reference video dataset, 2) a set of predefined attributes $\cal A$ is extracted, to 3) project each video into an attribute vector space that 4) is vector quantized. 5) the expected ABC of each cell is finally computed by measuring model success/failure rate for its videos. Bottom: to characterize complexity, an unseen video-question pair is first quantized in attribute space, and the expected ABC of the resulting cell returned as a complexity score.}
    \label{fig:teaser}
\end{figure*}

We argue that the problem has two main causes. The first is the reliance on {\it parametric} scores of complexity. Given the complexity of predicting video-LLM failures, it is hard to see how this can be done by a model smaller than the target video-LLM. One possibility is to build small parametric extensions that leverage features of the intermediate target video-LLM layers. However, these require at least partial processing of the video, which can  be expensive for larger models, and it is likely that the most informative features will be those of the later layers \citep{wiegreffe2024answer}, at which point there are not many computational savings. The second is the use of approaches that ignore explainability, \ie fail to define the vocabulary of video attributes that contribute to complexity and characterize complexity in terms of these attributes. To address these problems, we propose a {\it non-parametric} alternative based on the {\it explicit} characterization of complexity in terms of video attributes. We denote the approach as {\it Attribute-Based Complexity} (ABC) of video, or \ours. 

\ours is a non-parametric complexity score that, as illustrated in Figure~\ref{fig:teaser}, is computed with respect to an external reference video dataset $\cal D$, as is usual for {\it retrieval-augmented generation} (RAG) architectures. Given a dataset ${\cal D}$ of reference videos, a vocabulary of complexity attributes $\cal A$ is first defined and procedures designed to extract or control the values of the video attributes in $\cal A$. Each video is then represented as a vector in an attribute vector space ${\cal S}_{\cal A}$ whose dimensions are the attributes of $\cal A$. The space is then quantized, to produce a codebook $\cal C$ of representative attribute vectors. The expected ABC of each quantizer cell is then computed by processing its video-question pairs and tallying successes and failures. Given an unseen video, the associated attribute vector is computed and quantized with $\cal C$, and the complexity score of the associated bin is returned. 

\ours supports any attribute vocabulary $\cal A$. In this work, we use a vocabulary inspired by the historical evolution of video benchmarks. Early benchmarks~\citep{kay2017kinetics, soomro2012ucf101} typically consisted of short videos with relatively simple visual dynamics. As video understanding models improved, benchmarks increased the level of difficulty by introducing longer videos and more complex questions~\citep{li2018resound, zhou2024mlvu, xiao2021next}. Over time, as models adapted to these challenges, benchmarks~\citep{mangalam2023egoschema, wu2024longvideobench, liu2024tempcompass, rawal2024cinepile} further raised difficulty by incorporating even longer videos, more visually complex frames, and faster or denser events. This progression suggests the importance of attributes like video length, event location, event speed, and frame complexity. We thus use these attributes as attribute vocabulary $\cal A$. While we do not claim that this vocabulary is exhaustive, our results show that it enables effective predictions of video complexity with low-dimensional attribute vectors. 

The central technical component of \ours is the quantizer $\cal C$, which determines the balance of performance between in-distribution (ID) data (similar to $\cal D$) and out-of-distribution (OOD) data (unrelated videos). In order to enable the deployment of \ours without massive reference datasets, we propose a combination of two quantizers. The first is an in-distribution quantizer, learned by applying k-means to the attribute vectors extracted from $\cal D$. The second is a universal quantizer designed using synthetic videos with carefully controlled attribute values, so as to provide coverage of the entirety of the attribute space ${\cal S}_{\cal A}$. This is a lattice quantizer, where $\cal C$ is implemented as ${\cal C} = {\cal C}_1 \times \ldots \times {\cal C}_{|{\cal A}|}$ and ${\cal C}_k$ is a scalar quantizer for attribute $k$. This enables the training of the quantizer one-attribute-at-a-time, using a  dataset of synthetic video-question pairs ${\cal F}_k$ where only attribute $k$ is varied.   
For this, we propose a procedure inspired by psychophysics studies of human attention \citep{james1890principles, treisman1980feature, treisman1969strategies, treisman1986features, treisman1982illusory, treisman1991search, kahneman1992reviewing, treisman1960contextual, wolfe1989guided, chun1996just, wolfe1994guided, wolfe1998can}, based on  the manipulation of a target object embedded into a video of distractor objects. This procedure is used to  generate a dataset of synthetic video-question pairs   conditioned on user-specified attribute values, so as to cover the entire range of values of each complexity attribute. 

Experimental results show that \ours is significantly more effective at predicting the complexity of video-question pairs than various parametric approaches, including the popular use of an external video-LLM as judge. The gains are significant in terms of both the calibration of the complexity score and its computational efficiency. We also demonstrate how the explainable nature of a complexity score based on well-defined attributes provides insights on how the attribute composition of benchmarks affects their complexity, and the sensitivity of different models to different complexity attributes. In the supplement, we provide some preliminary examples of how \ours can be leveraged to implement effective curriculum learning and model distillation strategies. 

\noindent{\bf Contributions.} Our main contributions are as follows: 1) the \ours framework to quantify video-question complexity in terms of well-defined video attributes, 2) a procedure for effective attribute quantizer design from small reference datasets using 3) a combination of k-means and lattice quantization, and 4) a procedure for the synthesis of videos with controllable attributes to allow lattice-based complexity estimation. We also present experiments illustrating of \ours can provide valuable insights about video benchmarks and applications.




\section{Background}
\label{sec:related}

\noindent{\bf Video Attributes.} The study of video attributes that affect model performance has a long history in computer vision. For example, early work in representation bias~\citep{li2018resound} identified the tendency of early video datasets (\eg KTH~\citep{schuldt2004recognizing}, Weizmann~\citep{gorelick2007actions}, UCF101~\citep{soomro2012ucf101}, Kinetics~\citep{kay2017kinetics}) to favor short-term video representations. This motivated the introduction of datasets, like Diving48~\citep{li2018resound},  MVBench~\citep{li2024mvbench}, LongVideoBench~\citep{wu2024longvideobench}, MLVU~\citep{zhou2024mlvu}, ApoLLo~\citep{zohar2025apollo}, Cinepile~\citep{rawal2024cinepile}, NextQA~\citep{xiao2021next}, TempCompass~\citep{liu2024tempcompass} to name a few, without this property and work on long-term video representations, which are now the focus of most video model design efforts~\citep{zhang2025videollama, yang2025qwen2, li2024llava, lin2024video, laurenccon2024matters}. More recently, some work in the literature has started to address the problem of characterizing the attributes of video complexity. For example, the benchmark of \citep{mangalam2023egoschema} comprehensively evaluates how video-LLM performance depends on event length. All these works tend to show that attributes are primary factors affecting video complexity. Motivated by their insights, we propose a complexity score that depends only on these attributes.

\noindent{\bf Characterizing Complexity.} Earlier video benchmarks, such as those introduced in \citep{schuldt2004recognizing, gorelick2007actions, soomro2012ucf101}, primarily evaluated video understanding models on short clips. Subsequent datasets incorporated more visually complex scenes \citep{kay2017kinetics}. As more capable video-LLMs emerged \citep{yang2025qwen2, yang2025qwen3, li2024llava, lin2024video}, the community sought to increase benchmark difficulty by curating short videos with rapidly evolving events \citep{caba2015activitynet, kuehne2011hmdb, li2018resound}, \ie scenarios in which models must process a larger number of frames to answer correctly. With the advent of transformer-based architectures and, consequently, video-LLMs with substantially longer context windows \citep{yang2025qwen2, li2024llava}, this challenge was largely mitigated. More recent benchmarks~\citep{fu2025video, zhou2024mlvu, wu2024longvideobench, mangalam2023egoschema, rawal2024cinepile, li2024mvbench, liu2024tempcompass, xiao2021next} now feature significantly longer videos, rapidly changing events, and a broader diversity of question types, moving beyond the earlier focus on action recognition \citep{kay2017kinetics, li2018resound}. Notably, this progression reveals a consistent pattern in benchmark design: increasing difficulty by systematically modifying video attributes. The benchmarks have transitioned from short videos to visually complex scenes, then to fast-evolving events, and ultimately to substantially longer videos. Inspired by this trend of evolving benchmark curation, we propose to directly leverage these attribute dimensions as explicit dimensions of complexity.


\section{The \textbf{\texttt{VideoABC}} framework}
\label{sec:method}

In this section, we introduce {\it Video Attribute Based Complexity} (\ours), a non-parametric framework for characterizing video complexity in terms of video attributes. 

\subsection{Complexity measure}


\ours is a non-parametric measure of video-question pair complexity. Let $\mathbf{v} = [v_1, \cdots, v_{N}] \in \mathbb{R}^{N \times H \times W \times 3}$  be a video with $N$ RGB frames $v_k \in \mathbb{R}^{ H \times W \times 3}$ of height $H$ and width $W$. Given a video-LLM ${\cal M}$, the complexity posed by video-question pair $({\bf v},q)$ to $\cal M$ is a score $c \in [0,1]$, reflecting the probability that $\cal M$ will fail to solve $q$. 
Let ${\cal A}=\{a_1,\cdots,a_D\}$ be a set of $D$ video attributes and $\mathbf{a} = \alpha({\mathbf{v}, q})$ the embedding of $({\mathbf{v}, q})$ in a unit $D$-cube ${\cal S}_{\cal A}=[0,1]^{D} $ of attributes defined by $\cal A$.
\ours computes a nonparametric measure of complexity $c = \gamma(\mathbf{a}, {\cal D})$ using a reference dataset  ${\cal D} = \{({\bf v}_i, q_i)\}_i$ of video-question pairs, using the two-stage procedure of Figure~\ref{fig:teaser}. 

In a training stage, ${\cal D}$  is used to design a vector quantizer of ${\cal S}_{\cal A}$, consisting of a codebook ${\cal C}$ with attribute templates $\{\mathbf{c}_j\in {\cal S}_{\cal A}\}_{j=1}^{J}$ and associated quantization rule, 
\begin{equation}
    {\cal Q}(\mathbf{a}; {\cal D}) = \arg\min_{j\in\{1,\cdots,J\}} ||\mathbf{a} - \mathbf{c}_j ||,
    \label{eq:Q}
\end{equation}
which assigns $\mathbf{a}$ to the nearest neighbor template. Each attribute vector $\mathbf{a}$ in ${\cal D}$ is then mapped by $\cal Q$  into one of the quantizer cells
\begin{equation}
{\cal N}_j = \{\mathbf{a} | \,  ||\mathbf{a} - \mathbf{c}_j || \leq ||\mathbf{a} - \mathbf{c}_l ||, \forall l \neq j\}, j \in \{1, \ldots, J\}.
\end{equation} 
The model $\cal M$ is then applied to each of the pairs $({\bf v}_i, q_i) \in {\cal D}$ such that $\alpha({\bf v}_i) \in {\cal N}_j$ and a binary function  $y({\bf v},q)$ used to indicate success ($y=1$) or failure ($y=0$) in answering question $q_i$. The {\it Expected Attribute Based Complexity } (EABC) of cell ${\cal N}_j $ is finally defined as the failure rate of the videos mapped to the cell, 
\begin{equation}
    \psi_j = 1- \frac{1}{|\{i | \alpha({\bf v}_i, q_i) \in {\cal N}_j\}|} \sum_{i | \alpha({\bf v}_i, q_i) \in {\cal N}_j} y({\bf v}_i, q_i).
    \label{eq:psi}
\end{equation}
This procedure produces a pair of attribute template and video complexity  $(\mathbf{c}_j, \psi_j)$ per cell ${\cal N}_j$.

At inference, the complexity of a video question pair $({\bf v},q)$ is computed by quantizing $\bf v$ with \eqref{eq:Q} and returning the associated cell EABC, 
\begin{equation}
    \gamma({\bf v}, q; {\cal D}) = \psi_{{\cal Q}(\alpha({\bf v}, q); {\cal D})}.
    \label{eq:gamma}
\end{equation}

\subsection{Codebook Design}
\begin{figure*}[t]\RawFloats
  \centering
  \scriptsize
  \begin{minipage}{0.67\linewidth}
       \includegraphics[width=\textwidth]{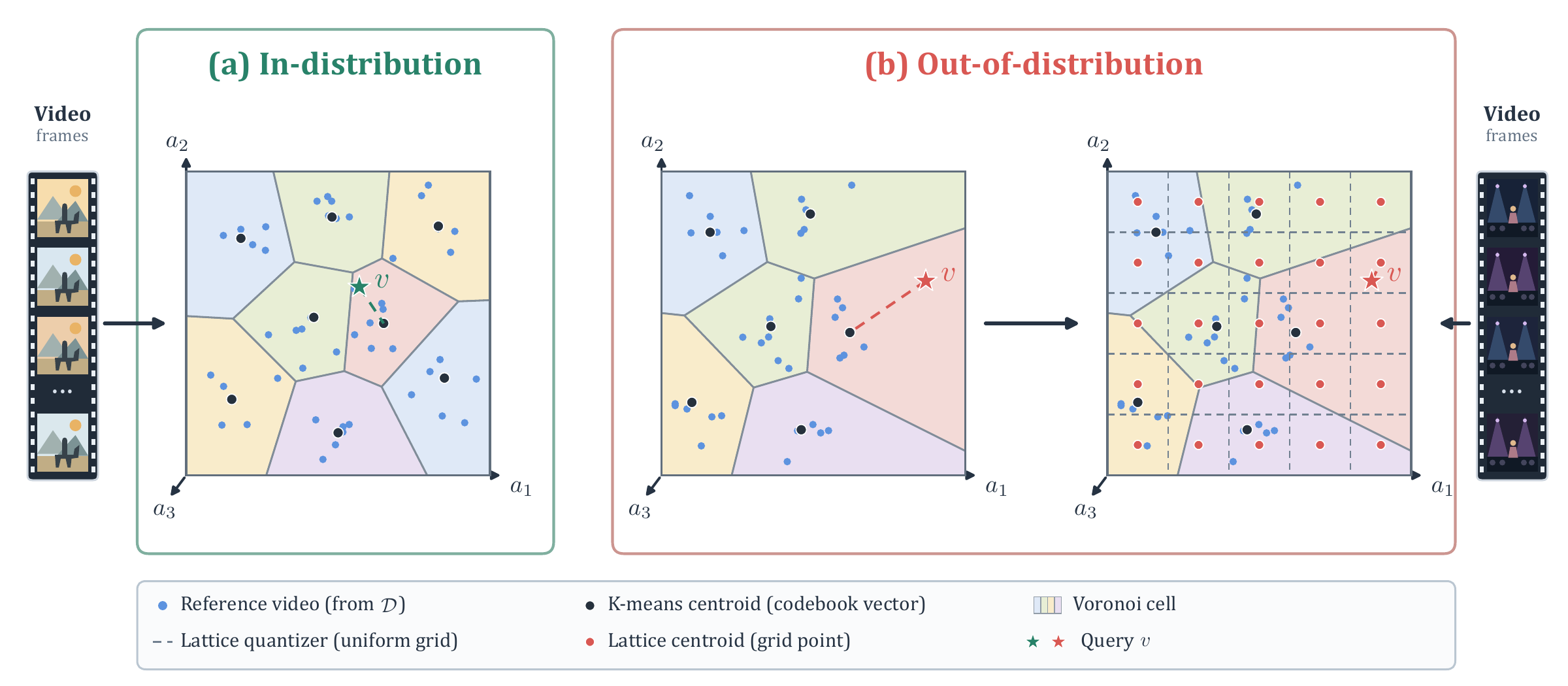} 
         \caption{Left: a K-means quantizer $\cal C$ is effective for videos similar to those of  $\cal D$ (in-distribution). Right: however, out-of distribution videos can land on large and sparsely populated cells, leading to poor ABC estimates. A universal lattice quantizer ${\cal C}^u$ provides uniform coverage of the space, improving generalization.}
    \label{fig:quantizers}
  \end{minipage}
  \,
  \begin{minipage}{0.30\linewidth}
  \centering
  \begin{tabular}{@{}c@{}}
     \includegraphics[width=\linewidth]{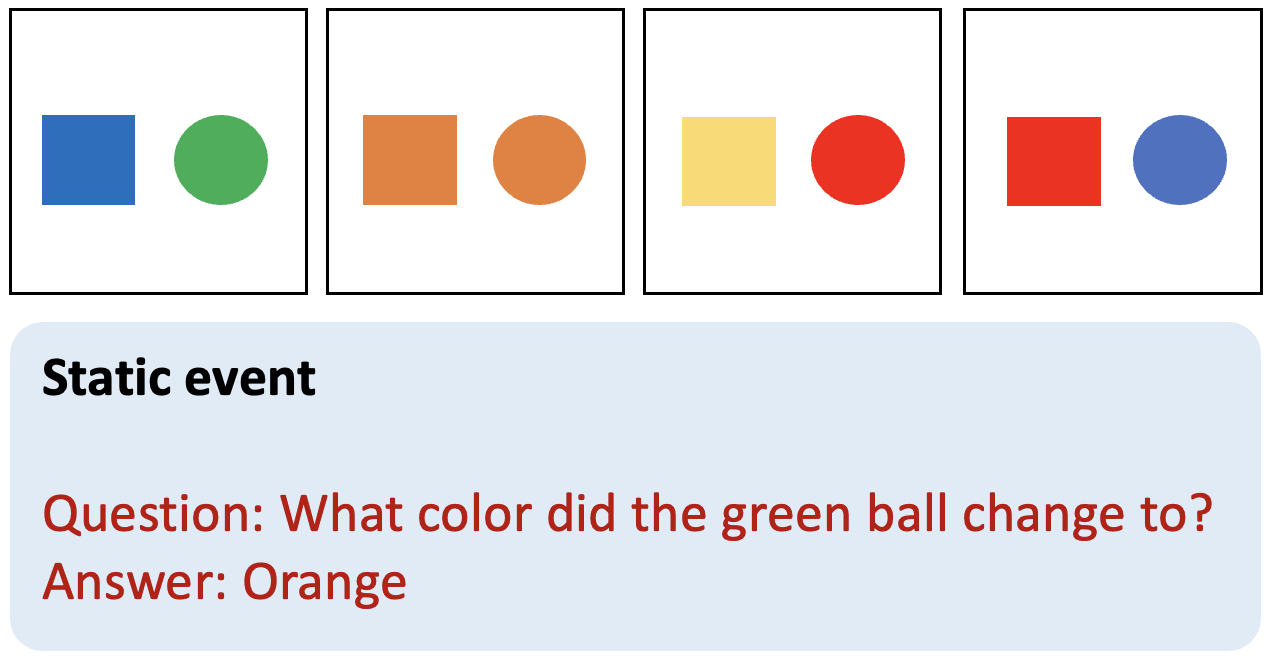}  \\
     \\
    \includegraphics[width=\linewidth]{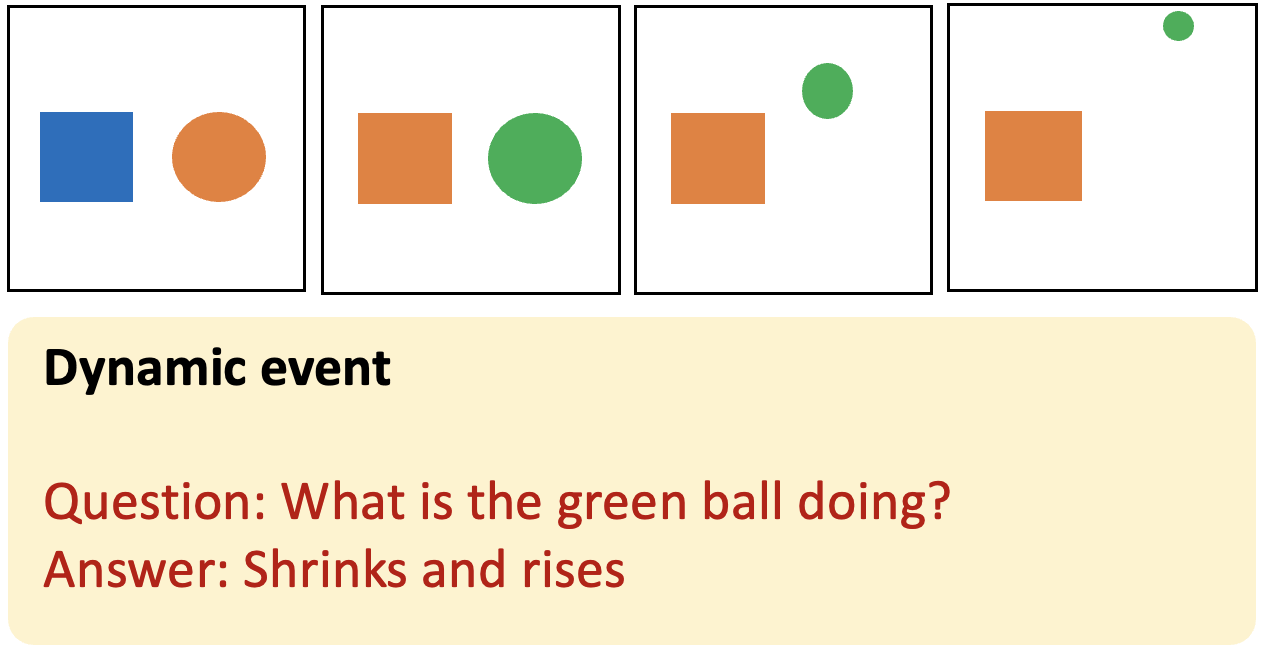} \\ \\
  \end{tabular}
    \caption{Examples of video-question pairs with static and dynamic events used to train the universal quantizer.}
    \label{fig:synthetic}
  \end{minipage}

\end{figure*}

The simplest strategy to design the codebook $\cal C$ is to apply a clustering algorithm to the reference dataset $\cal D$. In this work, we rely on k-means~\citep{Duda2000}. However, the accuracy of the resulting complexity measure of (\ref{eq:gamma}) depends on the relationship between $\bf v$ and  $\cal D$, as illustrated in Figure~\ref{fig:quantizers}. If $({\bf v},q)$ is in-distribution, it will likely be mapped to a densely populated populated cell, leading to an accurate estimate. However, if $({\bf v},q)$ is out of distribution, it could easily land on a large and sparsely populated cell. In this case, $\gamma({\bf v},q; {\cal D})$ can be a poor estimate of the real complexity. One strategy to address this problem is to rely on a very large $\cal D$. However, beyond the practical difficulties of learning with large datasets, it is difficult to know when $\cal D$ covers the space ${\cal S}_{\cal A}$ well enough to avoid out-of-distributions problems. In this work, we consider the alternative of relying on a {\it universal codebook} ${\cal C}^u$, implemented with lattice quantization~\citep{Conway1999}. This consists of designing ${\cal C}^u$ as a regular template grid, i.e.  ${\cal C}^u = {\cal C}^u_1 \times \ldots \times {\cal C}^u_D$, where ${\cal C}^u_k$ is a scalar quantizer of attribute $a_k$. Each dimension $k$ of ${\cal S}_{\cal A}$ is divided into $P$ uniform bins $\{ {\cal B}_{k,j} = [b_{j-1}=\frac{j-1}{P}, b_{j}=\frac{j}{P})\}_{j=1}^P$ of center point $c_{k,j} = \frac{2j-1}{2P}$, where the $b_i$ are bin delimiters shared by all the $D$ dimensions. This produces a total of  $P^D$ bins. The centers of these bins are used as templates of ${\cal C}^u$, \ie $\mathbf{c}_j=[c_{1,j}, \cdots, c_{P,j}]$ . 

While the use of this quantizer guarantees a uniform coverage of the attribute space by ${\cal S}_{\cal A}$ by the quantization cells ${\cal N}_k$, it makes the computation of the EABC of every cell difficult. Assume that $C$ videos are required per cell, to produce an accurate estimate of the EABC. Then, the computation of (\ref{eq:psi}) for cell ${\cal N}_k$ requires a dataset of $C$ videos with the combination of attribute values that make up the template attribute vector $\mathbf{c}_k$. However, identifying videos with this particular combination of attribute values in the wild is a `needle-in-the-haystack' problem. To overcome  this problem, we propose to use instead a reference dataset ${\cal D}^u$ of synthetic video, which allows the generation of videos with attribute vector $\mathbf{c}_k$ on demand. We then generate $C$ videos per cell, process them with the target video-LLM, compute success/failures and use (\ref{eq:psi}) to compute the cell EABC. This produces the set of EABCs $\psi^u_j$ associated with the templates $c_j$ of the universal quantizer ${\cal C}^u$.

However, even this procedure can be too complex, since it requires the generation of $C\times P^D$ videos and  their processing by the target video-LLM. As the dimension $D$ of the attribute embedding increases, the computation becomes unfeasible. We address this problem by learning attribute complexities independently. 
For each scalar quantizer ${\cal C}_k^u$ of ${\cal C}^u$, we use the video generation procedure to generate $C$ videos per quantization bin of the values of attribute $a_k$. We then compute the bin EABC using an attribute specific version of (\ref{eq:psi})
\begin{equation}
    \psi_{k,j}^u = 1- \frac{1}{|\{i | \alpha_k({\bf v}_i, q_i) \in {\cal B}_{k,j}\}|} \sum_{i | \alpha_k({\bf v}_i, q_i) \in {\cal B}_{k,j}} y({\bf v}_i,q_i).
    \label{eq:psiu}
\end{equation}
where ${\cal B}_{k,j}$ is the $j^{th}$ bin of the quantizer ${\cal C}_k^u$ of the $k^{th}$ attribute $\alpha_k({\bf v},q)$ of pair $({\bf v},q)$ and $\psi^u_{k,j}$ the associated EABC. This only requires the generation of $C \times P\times D$ videos and the computation of $P\times D$ EABCs. Finally, for each cell ${\cal N}$ of ${\cal C}^u$, we find the corresponding bins ${\cal B}_{k, j({\cal N})}$ for each attribute dimension $k$, and approximate the cell EABC by the largest of the bin EABCs,
\begin{equation}
    \hat{\psi}^u({\cal N}) = \max_k \psi^u_{k, j({\cal N})}.
    \label{eq:psiuEABC}
\end{equation}
This assumes that attributes have a disjunctive (logical OR) relation, e.g. a video can be complex because the event of interest is fast {\it or} the scene is complex. 
This procedure produces a set of approximate EABCs $\hat{\psi}^u_j$ associated with the templates $\mathbf{c}_j$ of the universal quantizer ${\cal C}^u$.

\subsection{Video Attributes}  
The aim of this work is not to produce an exhaustive list of all attributes that affect video complexity. While \ours is applicable to any choice of $\cal A$, we focus on a small set of \emph{event-based} attributes. An {\it event} $\mathcal{E}(q) \subseteq \mathcal{V}$ is a video clip informative of a question $q$. Let $\mathcal{E}(q) = [{e}_1, \cdots, {e}_{N_E}]$ be an event of  $N_E \leq N$ frames, where ${e}_{k} = {v}_{j+k-1},  k \in \{1,\cdots,N_E\}$. We consider six attributes. Attribute $a_1$ is the {\bf event location}, \ie $a_1 \coloneqq j$. The complement of $\mathcal{E}(q)$ in $\mathbf{v}$, \ie $\bar{\mathcal{E}}(q) = \mathbf{v} \setminus \mathcal{E}(q)$, is denoted as the {\it non-event} video. Attribute $a_2$ is the {\bf non-event proportion}, $a_2 \coloneqq |\bar{\mathcal{E}}(q)|/N$. Attribute $a_3$, denoted as {\bf scene complexity}, is the number of objects unrelated to the event present in the video. We provide below more details on how this number is determined. Finally, attribute $a_4$ is the {\bf event speed}. This is the smallest frame rate $f$\footnote{The frame rate of $\cal V$ is  $\rho = \frac{N}{T}$, where  $T$ is the temporal duration of $\mathcal{V}$ in seconds.} at which the video-LLM correctly answers question $q$ from the resampled video, i.e. $a_4 \coloneqq \min \{\rho:{u(\mathbf{v}^\rho,q) =1]}\}$, where $\mathbf{v}^\rho$ is the video resampled at frame rate $\rho$, and $u(\mathbf{v},q) = 1$ if $q$ is correctly answered for $\mathbf{v}$ and $0$ otherwise. These four attributes were chosen because they have been previously hypothesized to affect complexity by the video understanding community. For instance, non-event proportion and scene complexity have been previously leveraged to curate hard benchmarks in \citep{li2024mvbench, wu2024longvideobench, zhou2024mlvu, rawal2024cinepile}. Similarly, event speed has been utilized in benchmarks like \citep{li2018resound,zhou2024mlvu, li2024mvbench}. To test the robustness of \ours to the inclusion in $\cal A$ of attributes uninformative of video complexity we also considered two such attributes. Attribute $a_5$ is the \textbf{ratio between median and maximum pixel values} in the video \ie $a_5 \coloneqq \text{median}(\mathbf{v})/\text{max}(\mathbf{v})$. Attribute $a_6$ is the \textbf{ratio between mean and maximum pixel values} \ie $a_6 \coloneqq \text{mean}(\mathbf{v})/\text{max}(\mathbf{v})$.

\subsection{Datasets and Complexity Measures}



\textbf{In-distribution quantizer.} To train the in-distribution quantizer ${\cal C}$, we use a reference dataset $\cal D$ of 300 videos from each of four SOTA benchmarks: MV-Bench \citep{li2024mvbench}, MLVU \citep{zhou2024mlvu}, TempCompass \citep{liu2024tempcompass} and NextQA \citep{xiao2021next}. The resulting dataset has 1200 real video-question pairs. To quantify the attribute values of a video–question pair $({\bf v}, q)$, we define the question-relevant event as the set of video frames relevant to the question. Specifically, event localization step first uniformly samples 256 frames from the video and extracts DINO~\citep{caron2021emerging} features. Events are identified by thresholding inter-frame feature changes, and question relevance is estimated using SigLIP~\citep{zhai2023sigmoid} similarity between the question and the middle frame of each detected event. We consider six attributes estimated from this extracted event.
Given $\hat{\mathcal{E}}(q)$, the values of all attributes other than $a_3$ follow trivially from the attribute definitions above. To estimate the scene complexity attribute $a_3$, we computed its average compression ratio \ie $a_3 \coloneqq \frac{1}{K} \sum_{i=1}^K c_{i}$ where $c_{i}$ is the compression ratio of frame $i$. 
This procedure generates a dataset of 1200 $6$-dimensional attribute vectors, to which k-means~\citep{Duda2000} is applied to assign these attribute vectors to the cells of quantizer $\cal C$. The EABC of each quantizer cell is finally estimated with (\ref{eq:psi}).

\textbf{Universal quantizer.} As discussed above, the design of ${\cal C}^u$ is based on a dataset of synthetic videos that allows the manipulation of attribute values one-at-a-time. In this case, since the attribute values are known for all videos, there is no need for a procedure to estimate attribute values. To generate synthetic videos with specified attribute values, we propose a procedure inspired by the psychophysics literature~\citep{fechner1860elemente, craik2014effects, geng2014attentional, ebbinghaus1885gedachtnis, broadbent2013perception, loftus1974reconstruction, tversky1974judgment}, using 1) videos such as those shown in Figure~\ref{fig:synthetic}  and 2) the  target-distractor manipulations popular in studies of human attention~\citep{wolfe2004attributes}.
Every video uses a green ball as target object, the video-LLM is asked a question $q$ grounded on an event ${\cal E}(q)$, and presented with five answers. Events are defined by the change of a single target property, \eg color in Figure~\ref{fig:synthetic} (top).  Two types of events are defined\footnote{More details about the dataset curation are provided in Appendix~\ref{sec:data_curation}.}. 
\begin{enumerate}
    \item {\bf Static events.} Manipulations change a static target property. As shown in Figure~\ref{fig:synthetic} (top), event $\mathcal{E}$ is a color change of the green ball at a unique {\it event location} (frame pair) in the video. The task is to choose from one of five possible locations.
    \item {\bf Dynamic events.} Manipulations are dynamic, namely the combination of (upward or downward) ball motion and variation (increase or decrease) of ball size. Figure~\ref{fig:synthetic} (bottom) shows an example where the ball moves up and shrinks. The task is to choose from one of the four possible combinations of motion and shape change plus `none of the above'. 
\end{enumerate}
The remaining balls and squares in the video are distractors chosen randomly (restricting ball colors to non-green) per frame pair. The number of these objects is the scene complexity attribute $a_3$. 
Static events are used to quantify video complexity as a function event position ($a_1$), non-event proportion ($a_2$) and ratios ($a_5$ and $a_6$). Dynamic events are used to quantify complexity as a function of scene complexity ($a_3$) and event speed ($a_4$). To learn approximate EABCs, for each attribute $a_k$ and attribute bin ${\cal B}_{k,j}$, we generate $C$ synthetic video–question pairs and use (\ref{eq:psiu}) to estimate the bin EABC. The total number of samples needed to cover the  $P$ attribute bins of each scalar quantizer ${\cal C}^u_k$ is  $C \times P$. Given all individual attribute EABCs, we use (\ref{eq:psiuEABC}) to compute the approximate EABC $\hat{\psi}_j^u$ of each cell ${\cal N}_j$ of ${\cal C}^u$. The overall dataset ${\cal D}^u$ needed to learn all EABCs has size $C \times P \times D$. It is also possible to learn exact EABCs, by generating $C$ videos with  the attribute combination $c_k$ of each of the $P^D$ bins ${\cal N}_k$ of ${\cal C}^u$ and using (\ref{eq:psi}) to compute the corresponding EABC $\psi_k^u$. This requires $C \times P^D$ videos.

\textbf{Combined quantizer.} As discussed above, the two quantizers have different weaknesses. The in-distribution quantizer ${\cal C}$ cannot finely cover the entire vector space $\mathcal{S}_{\mathcal{A}}$. Hence, it generalizes poorly to videos outside the distribution of ${\cal D}$. The universal quantizer ${\cal C}^u$ eliminates this problem, but provides a weaker EABC estimate for in-distribution videos, due to both the reliance on lattice quantization and the approximation of (\ref{eq:psiu}). This suggests combining the two quantizers, using ${\cal C}$ for videos in the distribution of $\cal D$ and ${\cal C}^u$ for out-of-distribution videos. For this, we propose the following simple rule
\[
\gamma(\textbf{v}, q; \mathcal{D}) =
\begin{cases}
  \psi(\mathcal{N}_j) & \text{if} \quad  \exists{}\text{ }c_j\in\mathcal{C} \text { s.t. } \|\alpha({\bf v}, q) - c_j\|_2^2 \leq \tau \quad \\
  \psi^u(\mathcal{N}_j) & \text{otherwise}. 
\end{cases}
\]
 This reflects the intuition that if the attribute vector of $({\bf v},q)$ is too far from the closest template of the in-distribution quantizer $\cal C$, then we rely on the out-of-distribution quantizer. Otherwise, we rely on in-distribution quantizer. The threshold $\tau$ chosen is 2.4 and is obtained by cross-validation.

\paragraph{\bf Computational cost.} \ours is extremely efficient at inference, since the extraction of the attribute vector $\alpha({\bf v},q)$ and the quantization operations required by either (\ref{eq:psi}) or (\ref{eq:psiu}) use orders of magnitude less computation than computing video-LLM features for $\bf v$.  The main computational cost is in quantizer design during the training stage. This requires performing video-LLM inference on all videos of $\mathcal{D}$ or $\mathcal{D}^u$, in order to compute (\ref{eq:psi}) or (\ref{eq:psiu}), respectively, for each quantizer cell. This cost depends on the quantizer, as discussed above, ranging from $\mathcal{O}(CP)$ for the in-distribution quantizer to $\mathcal{O}(CP^D)$ for the universal quantizer with exact EABCs.

\section{Experiments}
\label{sec: final_experiments}
In this section, we report on various experiments performed to evaluate the \ours framework. 

\noindent{\bf Implementation details.} 
We considered 6 SOTA video-LLMs of different sizes: Qwen-3-VL (4B, 8B) \citep{bai2025qwen3}, LLaVA-OV (0.5B, 7B) \citep{li2024llava}, LLaVA-NeXT (7B) \citep{zhang2024llavanextvideo,liu2024llavanext} and Video-LLaVA (7B) \citep{lin2024video}. Due to space constraints, we present results for 5 models in the paper and the remaining in supplementary. Models were used with default parameters and initialized from \texttt{Huggingface}. Four$\times$NVIDIA-A6000 GPUs were used for implementation. All code and datasets will be shared upon paper publication.

\noindent{\bf Metrics.} Since the EABC is an estimate of the probability of video-LLM failure, we rely on the Expected Calibration Error (ECE)~\citep{boken2021appropriateness, nixon2019measuring} to test the effectiveness of \ours. This metric is commonly used to evaluate the calibration of probability estimates~\citep{nixon2019measuring, bella2010calibration, dormann2020calibration}.  

\noindent{\bf Baselines.} Unfortunately, most complexity scores used in the recent literature \citep{platanios2019competence, eyzaguirre2025understanding, graves2016adaptive} are closed-sourced, making direct comparisons impossible. Instead, we compare \ours to several baselines of comparable architecture. The primary baseline is the popular \textbf{video-LLM-as-judge} method (denoted \judge), where an external video-LLM is asked to estimate the complexity of the video-question pair $({\bf v},q)$. This approach is widely used in the literature \citep{eyzaguirre2025understanding, graves2016adaptive, agarwal2021evaluating}. In our experiments, we use LLaVA-OV (7B and 72B) models as external judge, unless otherwise specified. Note that when the target video-LLM is one of these models the approach is internal. The second baseline, denoted as \mlp, is an internal method. It uses a \textbf{2-layer MLP} to predict complexity from a feature vector extracted at the input of the penultimate layer embedding of the target video-LLM. The MLP is a binary classifier, trained to predict the target video-LLM  failures ($y=0$) and successes ($y=1$). The third baseline, denoted as \direct, bypasses the computation of the EABC. We simply extract the attribute vector from $({\bf v},q)$ and use the attribute values directly to create a complexity score. Similarly to (\ref{eq:psiuEABC}) we use the maximum attribute value  $\max(\alpha({\bf v},q))$ as complexity measure. The  fourth baseline, denoted as \textbf{\texttt{MLP-A}}, is an attribute  \textbf{MLP classifier}, which maps attribute values to target video-LLM success/failures. The final baseline,  denoted as \texttt{Random}, is a \textbf{random score generator} that uniformly samples a number in $[0,1]$. 

\noindent{\bf Evaluation datasets.} \ours is evaluated on two datasets. The first is an in-domain (\id) dataset, composed of 1200 videos samples from the four datasets used to curate the reference dataset ${\cal D}$ on which the in-distribution quantizer $\cal C$ is trained.  The second is an out-of-domain (\ood) dataset composed of 1200 videos from Egoschema \citep{mangalam2023egoschema} and LongVideoBench \citep{wu2024longvideobench}. The OOD dataset is used for all experiments unless otherwise stated. 
\subsection{Results.}    
    

In this section, we present the major results characterizing the performance of \ours. Detailed ablations of model parameters are presented in the supplement. 

\begin{figure*}[t]\RawFloats
    \centering
    \setlength{\tabcolsep}{3pt}
    \begin{tabular}{cccc}
        \includegraphics[width=0.23\textwidth]{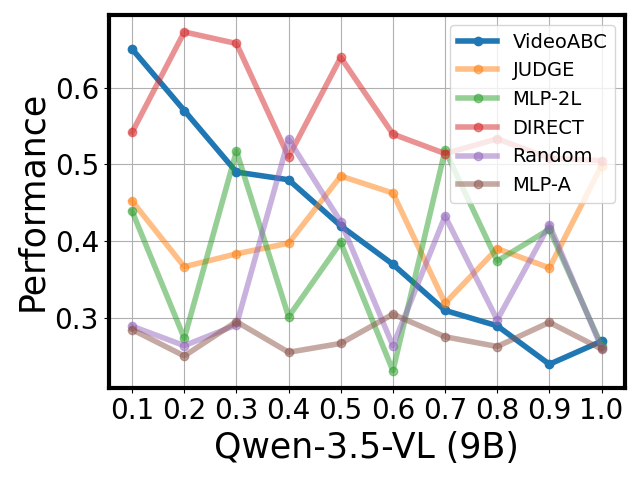} &
        \includegraphics[width=0.23\textwidth]{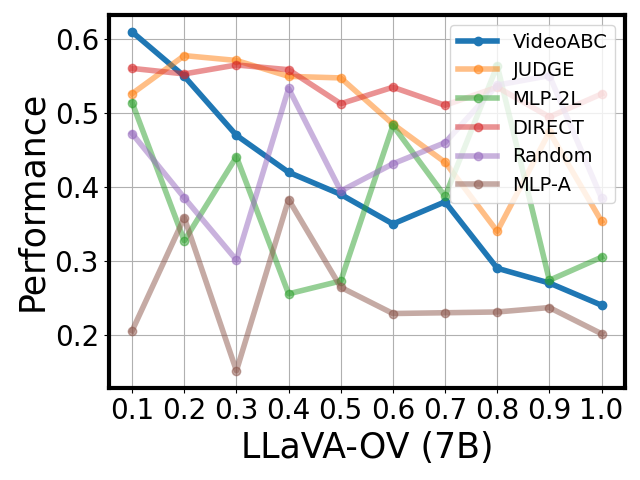} &
        \includegraphics[width=0.23\textwidth]{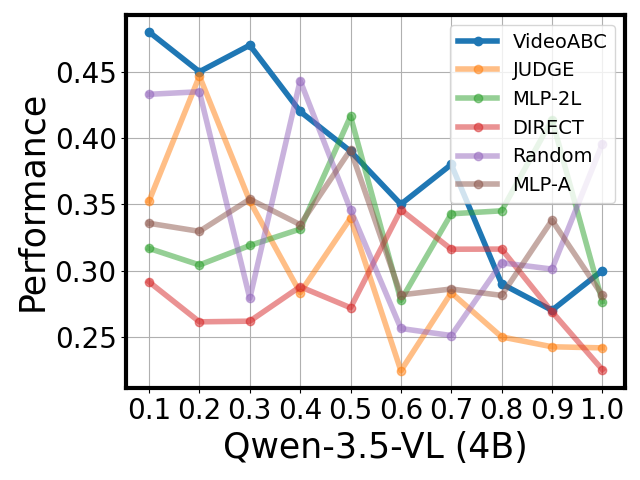} &
         \includegraphics[width=0.23\textwidth]{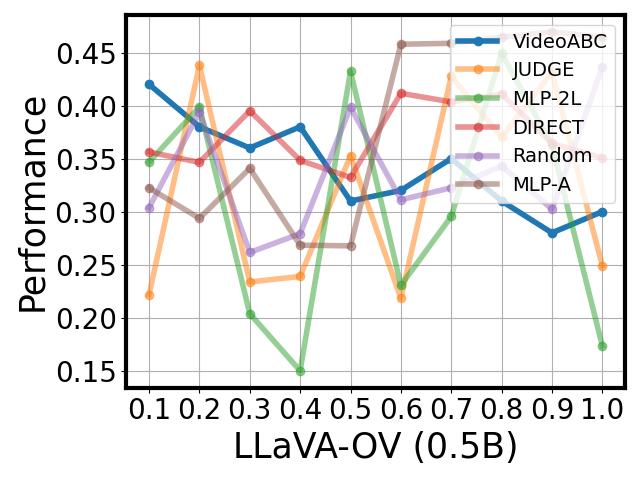} 
    \end{tabular}
    \caption{Video-LLM performance $p$ as a function of complexity score $\gamma$ for for Qwen-3-VL (4B, 8B) \citep{bai2025qwen3} and LLaVA-OV (0.5B, 7B) \citep{li2024llava}.}
    \label{fig: gen_plots}
\end{figure*}

\begin{figure}[t]\RawFloats
    \centering
    \begin{minipage}{.56\linewidth}
    \centering
    \scriptsize
    \setlength{\tabcolsep}{2pt}
    \begin{tabular}{l|ccccc|c}
    \toprule
    \multirow{1}{*}{Video-LLM} & \texttt{Random} & \judge &  \direct &  \mlp  & \textbf{\texttt{MLP-A}} & \ours\\
    \midrule
    & \multicolumn{6}{c}{Large size} \\
    \midrule
    Qwen-3-VL$_{(8B)}$ &  0.293 & 0.171 & 0.194 & 0.217 & 0.197 & \textbf{0.087} \\
    LLaVA-OV$_{(7B)}$ & 0.328 & 0.148 & 0.200 & 0.285 & 0.156 &  \textbf{0.058} \\
    Video-LLaVA$_{(7B)}$ & 0.346 & 0.173 & 0.189 & 0.292 & 0.192  &  \textbf{0.097} \\
    LLaVA-NeXT$_{(7B)}$ & 0.329 & 0.121 & 0.203 & 0.308 & 0.146 & \textbf{0.102} \\
    \midrule
    & \multicolumn{6}{c}{Small size} \\
    \midrule
    Qwen-3-VL$_{(4B)}$ & 0.348 & 0.208 & 0.246 & 0.322 & 0.198 &  \textbf{0.156} \\
    LLaVA-OV$_{(0.5B)}$ & 0.348 & 0.198 & 0.256 & 0.347 & 0.234 & \textbf{0.172} \\
    \bottomrule
    \end{tabular}
    \captionof{table}{Comparison of the ECE ($\downarrow$) of \ours and baselines for several target video-LLMs. LLaVA-OV (7B) was used as \judge.}
    \label{tab: ece_table}
    \end{minipage}
    \,\,\,
    \begin{minipage}{.39\linewidth}
    \centering
        \scriptsize
    \setlength{\tabcolsep}{2.5pt}
    \begin{tabular}{l|ccc}
    \toprule
       Method  & Size & Time (ms) & Performance \\
       \midrule
       \multirow{4}{*}{\judge}  & 0.5B & 167 & 0.187\\
        & 3B & 427 &  0.187\\
        & 7B & 656 & 0.171\\
        & 72B & 1802 & \underline{0.164}\\
        \midrule
        \direct & -- & 201 & 0.194\\
        \mlp & 8B & 678 & 0.217\\
        \textbf{\texttt{MLP-A}} & -- & \underline{119} & 0.182\\
        \midrule
        \texttt{Random} & - & \textbf{10} & 0.293\\
        \midrule
        \ours & -- & 226 & \textbf{0.087}\\
    \bottomrule
    \end{tabular}
         \captionof{table}{Comparison of model size, inference latency and ECE ($\downarrow$). }
         \label{tab: infer_effec}
    \end{minipage}
\end{figure}

\paragraph{\bf Qualitative results.} We start with a qualitative comparison between \ours and baselines in Figure~\ref{fig: gen_plots}. The figures show plots of video-LLM performance $p$ as a function of the video-question complexity score $\gamma$, for several video-LLMs. To produce these plots, we quantized the complexity score into ten bins and present the success rate per bin. The ideal plot would be a curve of the form $p=1-\gamma$ that also accounts for the fact that $p$ saturates and never reaches $0$ nor $1$. We did not attempt to compress or normalize the complexity scores to reflect this.The plots of \ours are much closer to ideal than those of all baselines. Most baseline curves lack the monotonically decreasing and almost linear trend of  \ours. The figure also shows that the difficulty of  complexity predicting is harder for the smaller models on the right. Beyond the compression of performance range, due to the lower model capacity that upper bounds the plots, most curves have higher $p$ values for the highest complexity ($\gamma=1$), suggesting that some easier video-question pairs receive high complexity scores.

\paragraph{\bf Quantitative results.} Table \ref{tab: ece_table} compares the ECE of the different methods for all six models. \ours achieves significantly lower ECE than all baselines, indicating stronger calibration with video-LLM performance, for all models. Among the baselines, \judge achieves the best performance. The internal \mlp method has surprisingly weak performance, showing that it may not be easy to extract signals predictive of success/failure from the target video-LLM itself. This is consistent with the well known difficulties of detecting LLM hallucinations~\citep{liu2024exploring, sriramanan2024llm}. The improved performance of \direct shows that video attributes are much more informative of video complexity and that the disjunctive assumption behind (\ref{eq:psiuEABC}) preserves much of that information. In fact, for the larger models \direct approaches the performance of \judge. \mlpa has similar but more variable performance. This is due to poor generalization of the MLP trained on a small set of real videos attribute vectors (only 1200). A more robust MLP would require a larger sample size and would significantly increase computational cost. Nevertheless, its gains over \direct for some target video-LLMs show that the disjuntictive assumption of (\ref{eq:psiuEABC}) can incurr losses. In conclusion, these results demonstrate that video attributes are reliable predictors of video-question complexity and that \ours produces significantly more reliable and efficient complexity estimates than existing baselines.


\paragraph{\bf Inference Latency.} Table~\ref{tab: infer_effec} compares the computational requirements of \ours and baselines\footnote{We exclude the training time required to construct the \ours codebook.}. \ours is approximately 8 times faster than a 72B \judge and has latency comparable to a 0.5B \judge. On the other hand, \ours substantially outperforms both the 72B and 0.5B \judge models. In other words, \ours matches the efficiency of a 0.5B \judge while outperforming even a 72B \judge. It also has no significant memory requirements. The latency of \ours is comparable to most other baselines, which \ours again outperforms substantially. The approaches with the next best trade-offs between complexity and perfrormance are \mlpa and \direct, again showing the benefits of ABC predictions.

\begin{table*}[t]\RawFloats
\centering
\scriptsize
\setlength{\tabcolsep}{3pt}
\begin{tabular}{l|c|cccccc||cccccc}
\toprule
\multirow{2}{*}{\ours} &  \multirow{2}{*}{Templates} & \multicolumn{6}{c}{\ood}  & \multicolumn{6}{|c}{\id} \\
& \cline{2-13} 
&  & $\mathcal{M}_1$ & $\mathcal{M}_2$ & $\mathcal{M}_3$ & $\mathcal{M}_4$ & $\mathcal{M}_5$ & $\mathcal{M}_6$ & $\mathcal{M}_1$ & $\mathcal{M}_2$ & $\mathcal{M}_3$ & $\mathcal{M}_4$ & $\mathcal{M}_5$ & $\mathcal{M}_6$ \\
\midrule
$({\cal C}, \psi)$ & 20 & 0.114 & 0.117 & 0.097 & 0.097 & 0.092 & 0.089 & \textbf{0.069} & \underline{0.062} & \underline{0.087} & \textbf{0.085} & \underline{0.074} & \underline{0.072} \\
$({\cal C}^u, \hat{\psi}^u)$ & 180 & \underline{0.082} & 0.062 & \underline{0.058} & \textbf{0.082} & 0.067 & \underline{0.059} & 0.095 & 0.088 & 0.104 & 0.096 & 0.092 & 0.092\\
$({\cal C}^u, \psi^u)$  & 10000 & \textbf{0.076} & \textbf{0.052} & \textbf{0.054} & \underline{0.087} & \underline{0.066} & \textbf{0.059} & 0.085 & 0.056 & \textbf{0.082} & \underline{0.087} & \textbf{0.068} &  \textbf{0.068} \\
\midrule
$({\cal C}, \psi) \cup ({\cal C}^u, \hat{\psi}^u)$  & 200 & 0.087 & \underline{0.058} & \textbf{0.054} & \textbf{0.082}  & \textbf{0.064} & 0.059 &  \underline{0.072} & \textbf{0.054} & 0.094 & 0.096 & 0.087 & 0.084 \\
\bottomrule
\end{tabular}
\caption{ECE and number of template computations for different EABC estimation strategies on the \ood and \id datasets. \textbf{Bold letters} denote best performance while \underline{underlined letters} denote second best performance. $\mathcal{M}$ denotes different video-LLMs in order described in implementation details (sec \ref{sec: final_experiments})}
\label{tab: id-ood}
\end{table*}

\begin{figure}[t]\RawFloats
    \centering
    \includegraphics[width=0.5\linewidth]{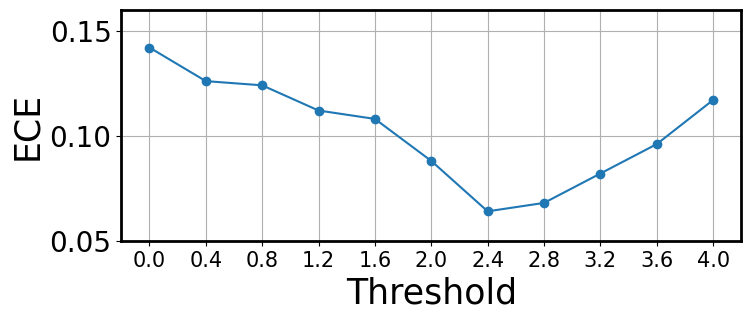}
    \caption{Figure show how ECE varies with respect to threshold of combined quantizer.}
    \label{fig: thresh}
\end{figure}

\begin{table}[t]\RawFloats
    \centering
    \scriptsize
    \setlength{\tabcolsep}{1.25pt}
    \begin{tabular}{cc||cc}
    \begin{tabular}{c}
    \includegraphics[width=0.23\textwidth]{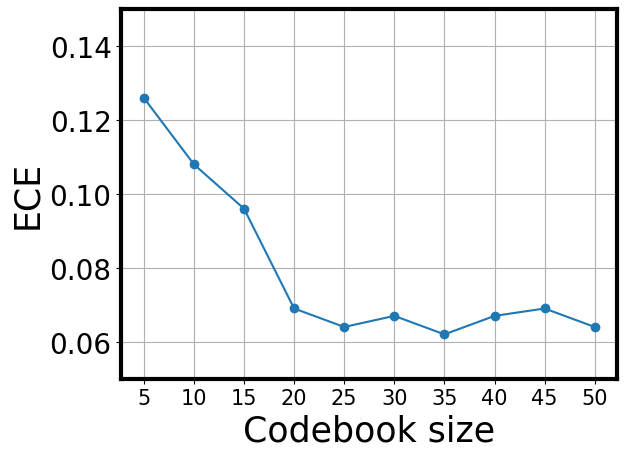}
    \end{tabular}
    & 
    \begin{tabular}{c}
    \includegraphics[width=0.23\textwidth]{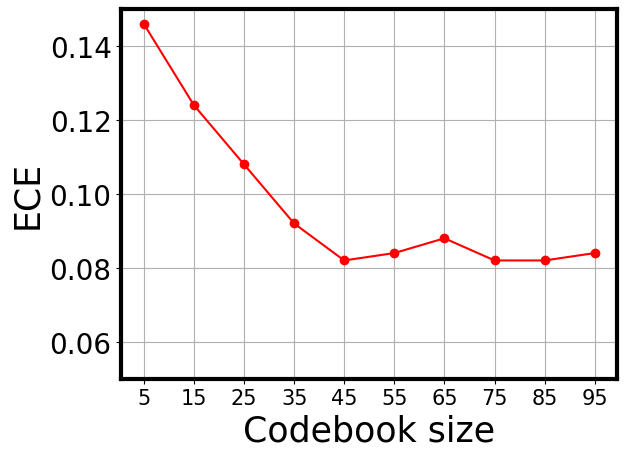}
    \end{tabular}
    & 
    \begin{tabular}{c}
    \includegraphics[width=0.23\textwidth]{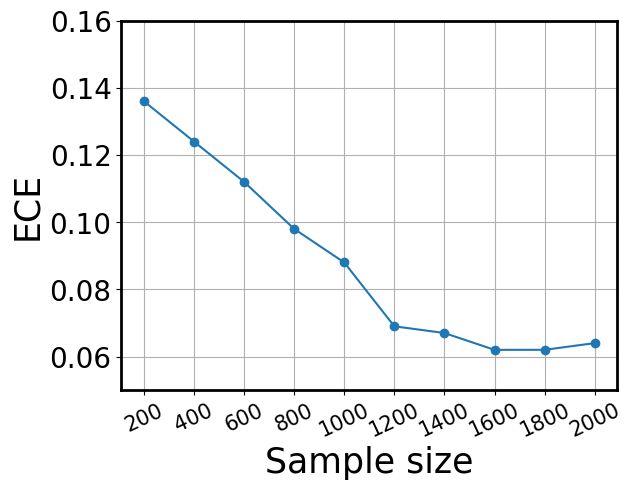}
    \end{tabular}
    & 
    \begin{tabular}{c}
    \includegraphics[width=0.23\textwidth]{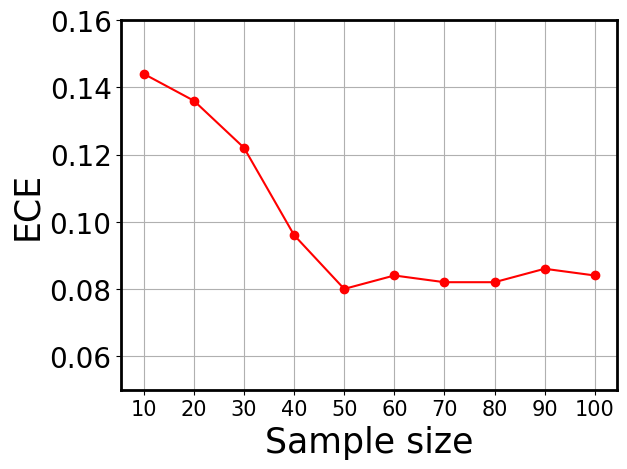}
    \end{tabular}
    \end{tabular}
    \captionof{figure}{Left: ECE ($\downarrow$) vs codebook size for in-distribution (blue) and universal (red) quantizers. Right: ECE ($\downarrow$) vs sample size for in-distribution (blue) and universal (red) quantizers. The in-distribution and universal quantizers were tested on \id and \ood datasets respectively.}
    \label{fig: abla_size}
\end{table}

\begin{figure*}[t]\RawFloats
    \centering
    \begin{minipage}{0.45\linewidth}
        \centering
        \includegraphics[width=.75\linewidth]{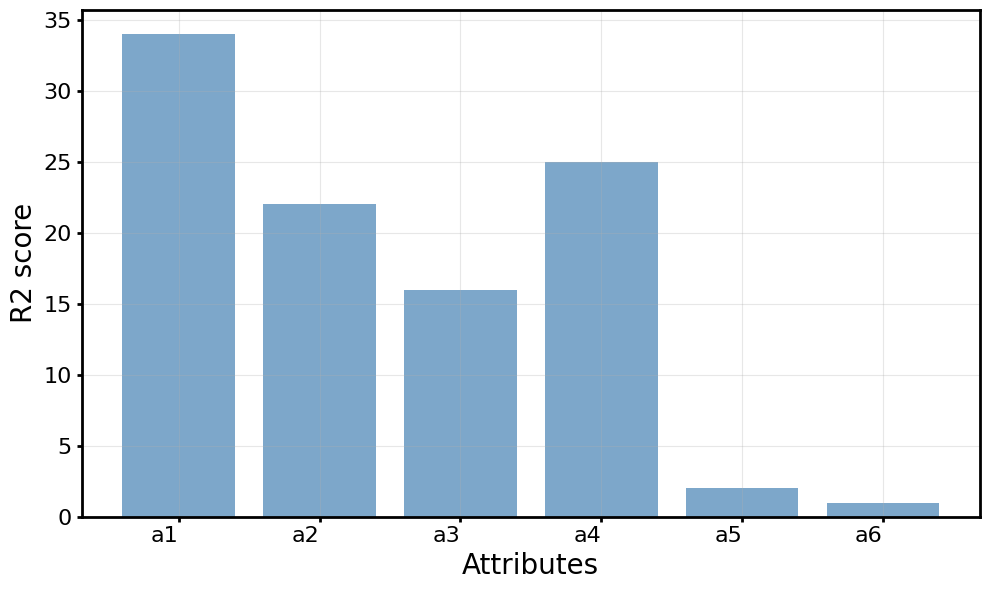}
        \caption{Attribute weights of \ours score regression.}
        \label{fig:regression}
    \end{minipage}
    \,
    \begin{minipage}{0.45\linewidth}
        \centering
        \begin{tabular}{c}
        \includegraphics[width=0.75\linewidth]{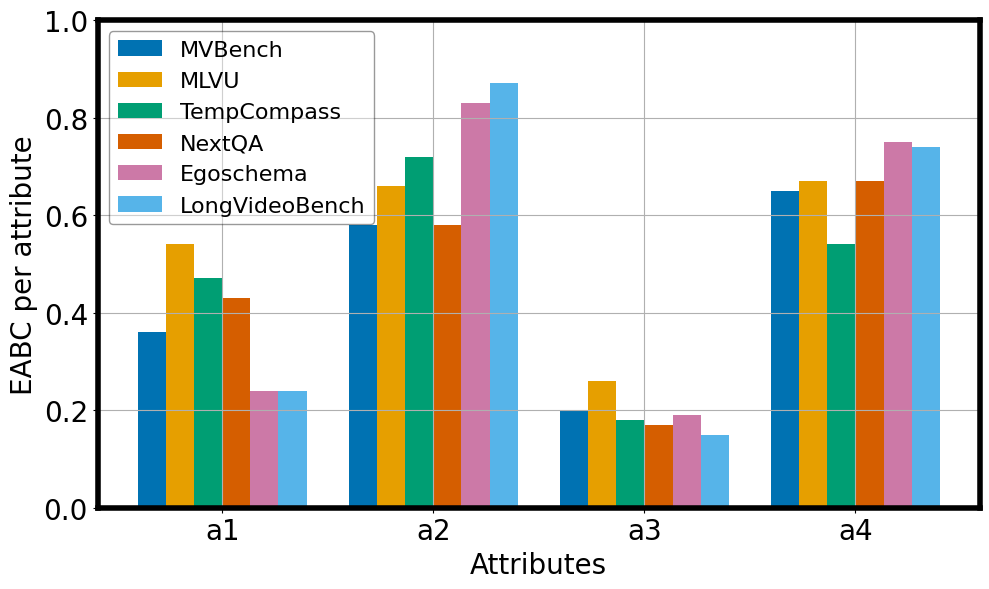} \\
        \end{tabular}      
        \caption{Average attribute EABC for different benchmarks. Attributes a$_5$ and a$_6$ are omitted, due to their minimal contribution to video complexity. }
        \label{fig:attECE}
    \end{minipage}
   \label{fig: effiency}
\end{figure*}

\paragraph{\bf Robustness.} To test the robustness of \ours to attribute noise, we included in $\cal A$ non-informative attributes $a_5, a_6$, which are  expected not to affect video-LLM performance. We performed a linear regression with attribute values as inputs and \ours scores as output. The magnitude of the weight learned per attribute measures its importance to the regression.  Figure \ref{fig:regression} shows these weights. The contributions of $a_5$ and $a_6$ are minimal compared to those of the remaining attributes. This shows that \ours can effectively ignore non-informative attributes. 

\subsection{Ablation}
\label{sec: ablate}
In this section, we discuss various ablations of \ours. 

\paragraph{\bf Codebook.} Table~\ref{tab: id-ood} compares variants of \ours with different quantizer and EABC computations: in-distribution quantizer $\cal C$, universal quantizer ${\cal C}^u$ with approximate EABC $\hat{\psi}^u$ or true EABC $\psi^u$ and combined quantizer $({\cal C},\psi) \cup ({\cal C}^u, \hat{\psi}^u)$.  Several observations can be  made. First, while the in-distribution quantizer $({\cal C},\psi)$ outperforms the approximate universal quantizer $({\cal C}^u, \hat{\psi}^u)$ for in-domain evaluation (\id), the latter has significantly better performance in the out-of-domain setting (\ood). This confirms that, while k-means is more effective for in-distribution data, lattice quantization guarantees better generalization. Second, while the universal quantizer with exact EABC $({\cal C}^u, \psi^u)$ has the overall best performance, it requires a large number of bins. This drastically increases the computation and storage needed to create videos, process them with the target video-LLM, and store them. However, it only sightly  outperforms the combined quantizer $({\cal C},\psi) \cup ({\cal C}^u, \hat{\psi}^u)$, which requires orders of magnitude less bins. Overall, the combined quantizer achieves the best trade-off between performance and training complexity. In fact, the implementation of \ours with the combined quantizer is close to the best performance for both the in-distribution and out-of-distribution settings, for all video-LLMs. We thus use this configuration in all experiments.

\paragraph{\bf Codebook size.} The sizes of both codebooks were varied in the experiments. Figure \ref{fig: abla_size} presents the results for the two codebooks across different sizes. Both in-distribution quantizer and universal quantizer was tested on concatenated \id and \ood datasets. Two key conclusions emerge. First, varying J in the in-distribution quantizer has little effect on performance beyond a certain point; in our experiments, this threshold was around 20. Beyond this value, there is no significant improvement in the model’s ECE. Second, for the universal quantizer, increasing P leads to improved performance initially, but gains diminish after an upper threshold, where performance begins to plateau. Overall, for both quantizers, performance improves with increasing values of $J$ and $P$ at lower ranges, followed by saturation. 



\paragraph{\bf Sample size.} Figure~\ref{fig: abla_size} (right) shows the effect of sample size on \ours performance. For the in-distribution quantizer ${\cal C}$, sample size is the size of the reference dataset $|{\cal D}|$. For the universal quantizer ${\cal C}^u$, it is the number of videos synthesize per attribute to estimate the per-attribute complexity of (\ref{eq:psiu}). One can note that as the total number of real videos increases, the computational complexity of the in-distribution quantizer gradually saturates to a fixed value. For the universal quantizer, performance improves as the sample size \emph{per bin} increases. For real videos, we report results in terms of total samples, whereas for synthetic data we report per-bin sample sizes. This distinction arises because controlling the number of samples per bin is straightforward in the synthetic setting, while for in-distribution data it is significantly more difficult, as it depends heavily on the available dataset. Our experiments further show that randomly sampling videos from diverse datasets effectively alleviates this limitation and yields stable performance.

\paragraph{\bf Threshold of combined quantizer.} 
We ablated the threshold of the combined quantizer, ($\tau$), on a concatenated \ood and \id dataset. The results are shown in figure \ref{fig: thresh}. As the threshold distance increases, performance initially declines slightly and then drops sharply. This behavior arises because predictions are increasingly dominated by the in-distribution quantizer, which is not well-suited for handling \ood samples. Conversely, when the threshold is reduced, most predictions are made by the universal quantizer; due to its approximate nature, its overall performance is also suboptimal. Notably, the performance degradation across threshold values is relatively modest, indicating that \ours is robust and does not degrade abruptly with changes in ($\tau$).

Applications of \ours is presented in Appendix.

\begingroup
\titlespacing*{\section}{0pt}{1.5ex}{1ex}
\section{Discussion}
\label{sec:conclude} 
\noindent{\bf Conclusion.} We have introduced \ours, a framework for characterizing video-LLM complexity with respect to a pre-defined attribute vocabulary. \ours is a non-parametric complexity measure, based on the quantization of the attribute space and a reference dataset combining real and synthetic videos. Experiments have shown that it produces complexity estimates significantly better calibrated than those of various baselines, including the popular \judge approach, with much less complexity. A preliminary analysis of the attribute complexity of several benchmarks illustrates  the potential of the measure for benchmark characterization.

\noindent{\bf Limitations.}The main limitation is that this work only considers four complexity attributes. We believe that the space of the latter is much higher dimensional. In this sense, the current version of \ours is simply a starting point for additional research. We have also only used \ours to perform a preliminary characterization of benchmark complexity, as the main goal of the work was to introduce and test the \ours measure itself. We leave these extensions for future exploration.

  \endgroup

\clearpage
\bibliographystyle{assets/plainnat}
\bibliography{main}

\clearpage
\beginappendix



\section{Applications}
\label{sec: bench_appl}

\paragraph{\bf Benchmark Composition.} 
 Figure~\ref{fig:attECE} presents average EABC per attribute of six widely used benchmark datasets, showing how different benchmarks pose challenges along different attribute dimensions. For example, LongVideoBench is very challenging in terms of non-event proportion (due to the long videos) but not in terms of event location or scene complexity. This finding suggests that adding more challenging videos along these attribute dimensions would lead to an overall more challenging benchmark. 
 More broadly, the figure shows some interesting observations about the benchmark landscape. First, no benchmark is most challenging across all attribute directions. MLVU has the most challenging event locations and scene complexities, while LongVideoBench dominates the non-event  proportion attribute, and Egoschema the event speed attribute. 
 Second, most benchmarks are much less challenging than the rest of the pack along at least one attribute. MLVU stands out as an exception, exhibiting a more balanced distribution, with its complexity stemming from all considered attributes. Finally, current benchmarks tend to emphasize videos that are complex in terms of non-even proportion and event speed. However, the coverage of challenging event locations and scenes is much weaker. 
This highlights the need for benchmarks that better capture diversity across these overlooked dimensions.

\begin{table}[b]\RawFloats
    \centering
    \scriptsize
    \setlength{\tabcolsep}{3pt}
    \begin{tabular}{cc}
    \begin{tabular}{l|ccccc}
        \toprule
        Video-LLM & Level 1 & Level 2 & Level 3 & Level 4 & Level 5\\
        \midrule
        Random & 0.25 & 0.25 & 0.25 & 0.25 & 0.25 \\
        \midrule
        & \multicolumn{5}{c}{Medium size} \\
        \midrule
        Qwen-3-VL$_{(8B)}$  & 0.69 & 0.58 & 0.53 & 0.47 & \textbf{0.36} \\
        LLaVA-OV$_{(7B)}$  & 0.72 & 0.64 & 0.67 & 0.49 & \textbf{0.41} \\
        Video-LLaVA$_{(7B)}$  & 0.65 & 0.59 & 0.51 & 0.39 & \textbf{0.32}\\
        LLaVA-NeXT$_{(7B)}$  & 0.66 & 0.53 & 0.48 & 0.36 & \textbf{0.29}\\
        \midrule
        & \multicolumn{5}{c}{Small size} \\
        \midrule
        Qwen-3-VL$_{(4B)}$ & 0.58 & 0.52 & 0.41 & 0.43& \textbf{0.24} \\
        LLaVA-OV$_{(0.5B)}$ & 0.60 & 0.58 & 0.38 & 0.36 & \textbf{0.26}\\
        \bottomrule
    \end{tabular} 
    &
    \begin{tabular}{c}
     \setlength{\tabcolsep}{3pt}
    \begin{tabular}{l|cc}
        \toprule
        Training type & LLaVA-OV$_{(0.5B)}$ & Qwen-3-VL$_{(4B)}$  \\
        \midrule
         Untrained & 0.43 & 0.57\\ 
         Random & 0.49 & 0.59\\ 
         \midrule
         \judge & 0.53 & 0.47\\ 
         \mlpa & 0.49 & 0.56\\
         \mlp & 0.47 & 0.59\\
         \direct & 0.53 & 0.61\\
         \midrule 
         \ours & \textbf{0.57} & \textbf{0.63}\\
        \bottomrule
    \end{tabular}
\end{tabular}
    \end{tabular}
    \caption{(Left) Performance of a LLaVA-OV (7B) model on real datasets with 5 levels. A random chance can secure an accuracy of 0.25 in both cases. (Right) Performance of a LLaVA-OV (0.5B) model on \ood. Model is trained on Next-QA training set.}
    \label{tab: curriculum}
\end{table} 

\paragraph{\bf Levels of benchmarks.} With the rapid advancement of video-LLMs, a growing number of benchmarks have been introduced, making comprehensive evaluation increasingly computationally expensive \cite{zohar2025apollo}. We show that \ours can effectively address this challenge by distilling large benchmarks into structured levels of difficulty. In this work, we construct five such levels, each comprising 200 samples, with the objective of maximizing the presence of videos that are simultaneously complex across all attributes. Specifically, we threshold \ours scores and curate a compact subset of 200 video samples for each level. To evaluate the effectiveness of this approach, we benchmark several video-LLMs on these distilled subsets. As shown in Table~\ref{tab: curriculum} (right), model performance consistently declines as the benchmark level increases, indicating that higher levels indeed correspond to more challenging samples. In conclusion, our results demonstrate that \ours enables efficient and effective benchmark distillation while preserving meaningful gradients of difficulty.

\paragraph{\bf Curriculum training.} We consider performing supervised fine-tuning (SFT) using a curriculum defined by \ours. Simply we train the model first on easier questions and then on harder questions defined by different complexity score baselines. We trained LLaVA-OV (0.5B) and Qwen-3-VL (4B) on training set of Next-QA and measured model performance on level 5 of our curated dataset. Results are shown in table \ref{tab: curriculum} (right). It can noted that both methods trained on our method shows much better results as compared to other baselines implying our curriculum is better as compared to standard one. 



\begin{figure}[t]\RawFloats
\centering
\begin{tabular}{ccc}
    \includegraphics[width=0.3\textwidth]{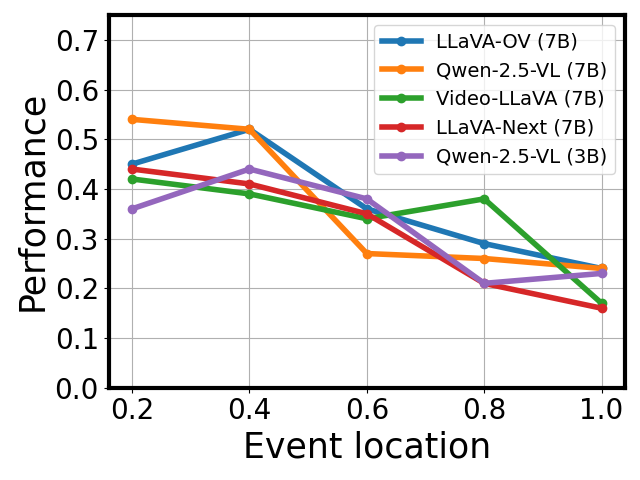} &
    \includegraphics[width=0.3\textwidth]{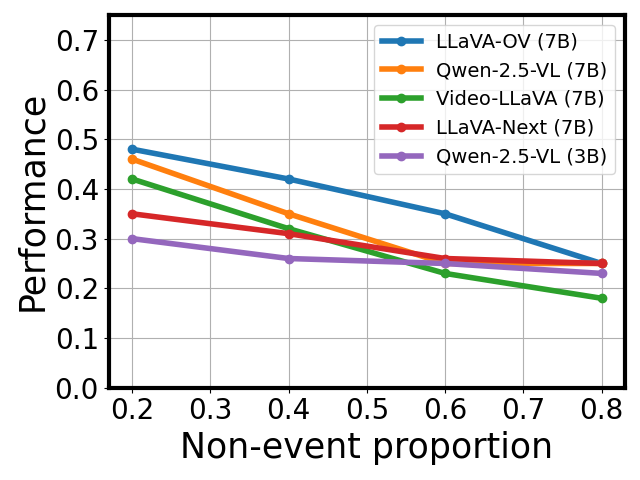} &
    \includegraphics[width=0.3\textwidth]{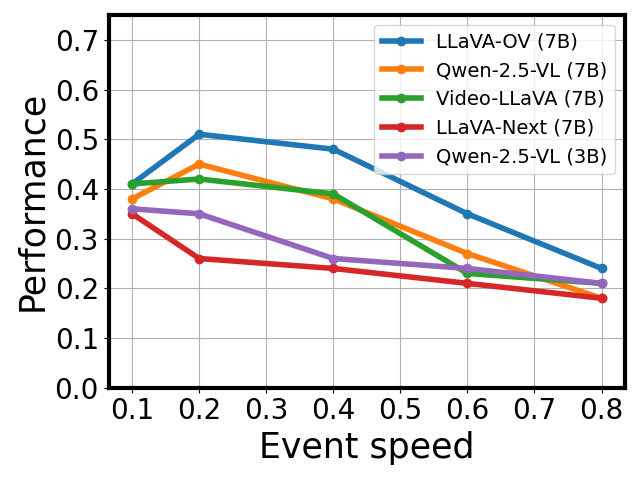} \\
    \includegraphics[width=0.3\textwidth]{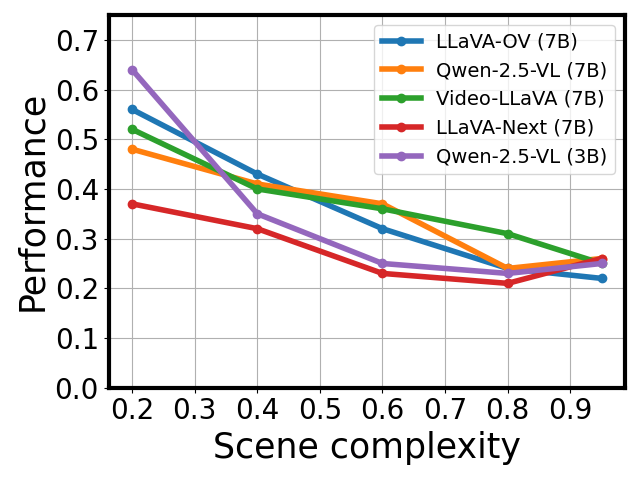} &
    \includegraphics[width=0.3\textwidth]{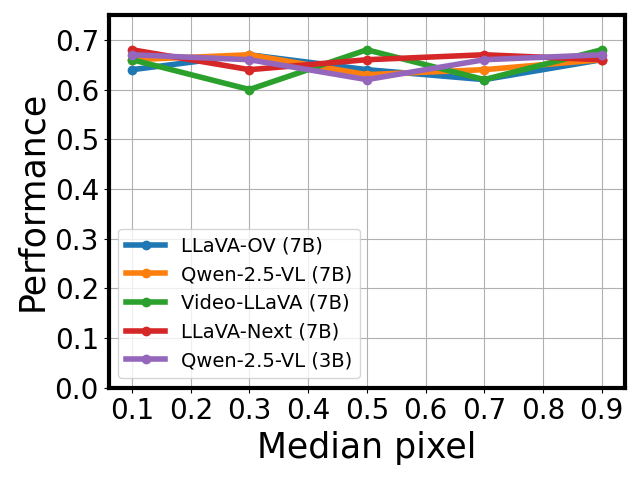} &
    \includegraphics[width=0.3\textwidth]{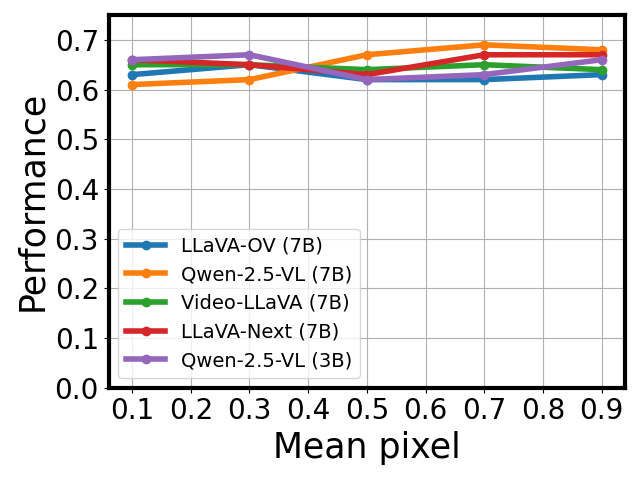}
\end{tabular}
\caption{Figure shows sensitivity of video-LLM.}
\label{fig: chloe-ben-effect}
\end{figure}

\paragraph{\bf Video-LLM sensitivity.} A complementary question is how sensitive is video-LLM performance to complexity along the various attribute dimensions. To investigate this, we leveraged the universal quantizer to measure model performance $p$ as a function of the complexity of each attribute $a$, while maintaining the remaining constant. Given the values $c_j$ of the attribute bin centers and the associated EABCs average model performance $p_j = 1- \psi^u_j$, the model sensitivity was measured with 
$s_j = (p_{j+1} - p_j)/(c_{j+1}-c_j)$. Figure \ref{fig: chloe-ben-effect} shows the model sensitivity of video-LLMs towards different attributes. Two conclusions can be made. To begin with, event speed attribute breaks most of video-LLMs at a lower complexity value while for other attributes, it is at some higher complexity value. The main reason for such a behavior is the lack of essential frames to answer the question. Event speed and scene complexity are primary factors and that is because the model is unable to get relevant frames to answer questions. Overall, the most effective attribute at breaking video-LLMs is event speed followed by scene complexity. Furthermore, it can be noted that model size is another contributing factor preventing model's breakage. However, it is effective for only non-event proportion and frame difficulty. For other attributes such as event speed, model size does not seem to be a solution. Overall, it shows that certain attributes are more effective at breaking video-LLM and there is a need to have video-LLM architectures that can resolve these.

\section{Data curation for universal quantizer}
\label{sec:data_curation}
\begin{figure*}\RawFloats
  \centering
  \scriptsize
  \begin{minipage}{0.3\textwidth}
    \centering
    \includegraphics[width=\textwidth]{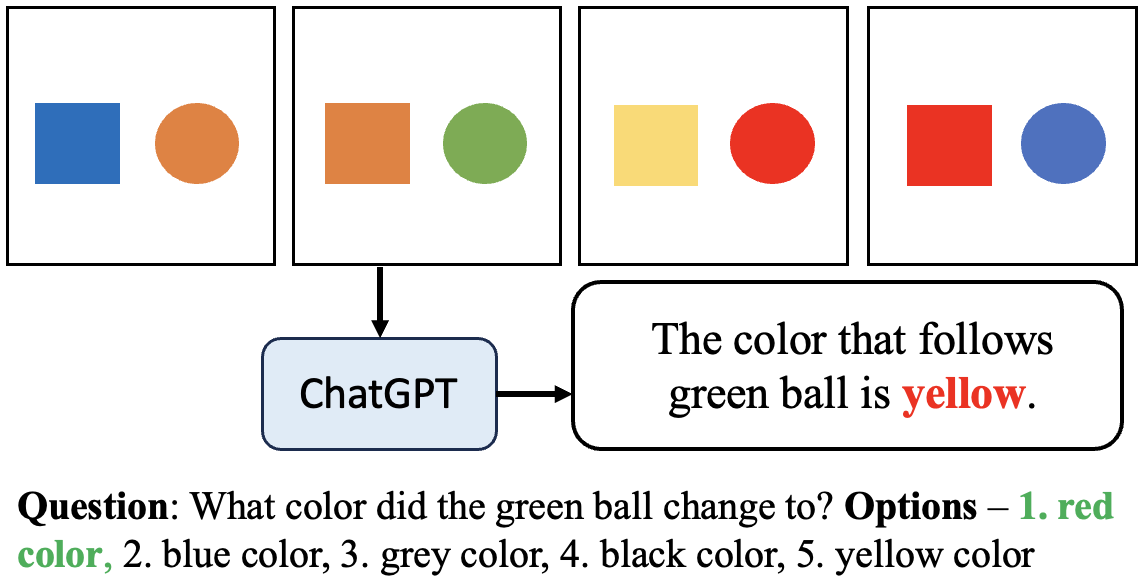}
    \subcaption{Static event: target manipulation.}
  \end{minipage}
  \hfill
  \begin{minipage}{0.3\textwidth}
    \centering
    \includegraphics[width=\textwidth]{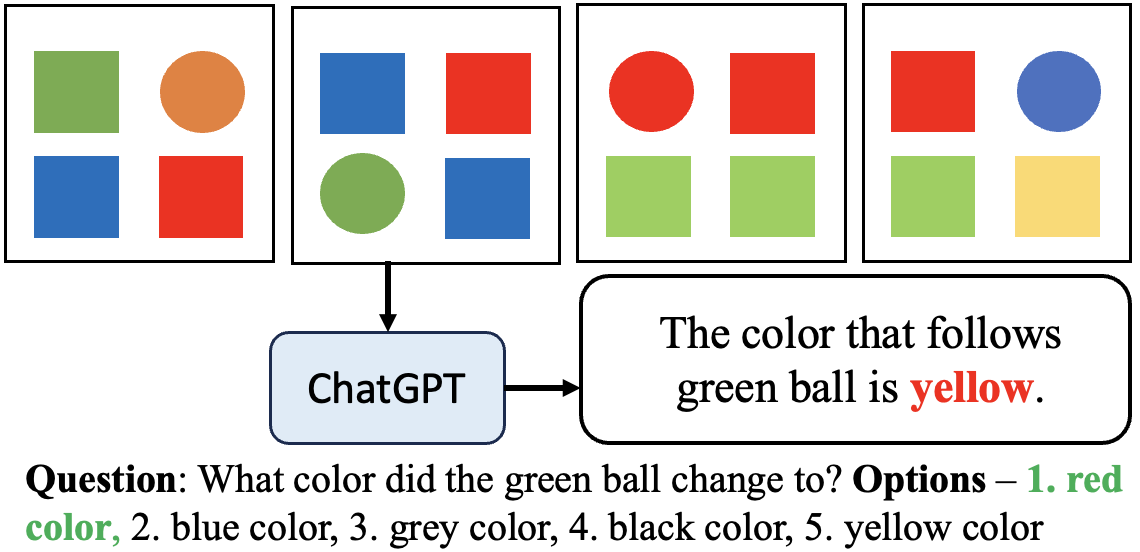}
    \subcaption{Dynamic event: background manipulation.}
  \end{minipage}
  \hfill
  \begin{minipage}{0.35\textwidth}
    \centering
    \includegraphics[width=\textwidth]{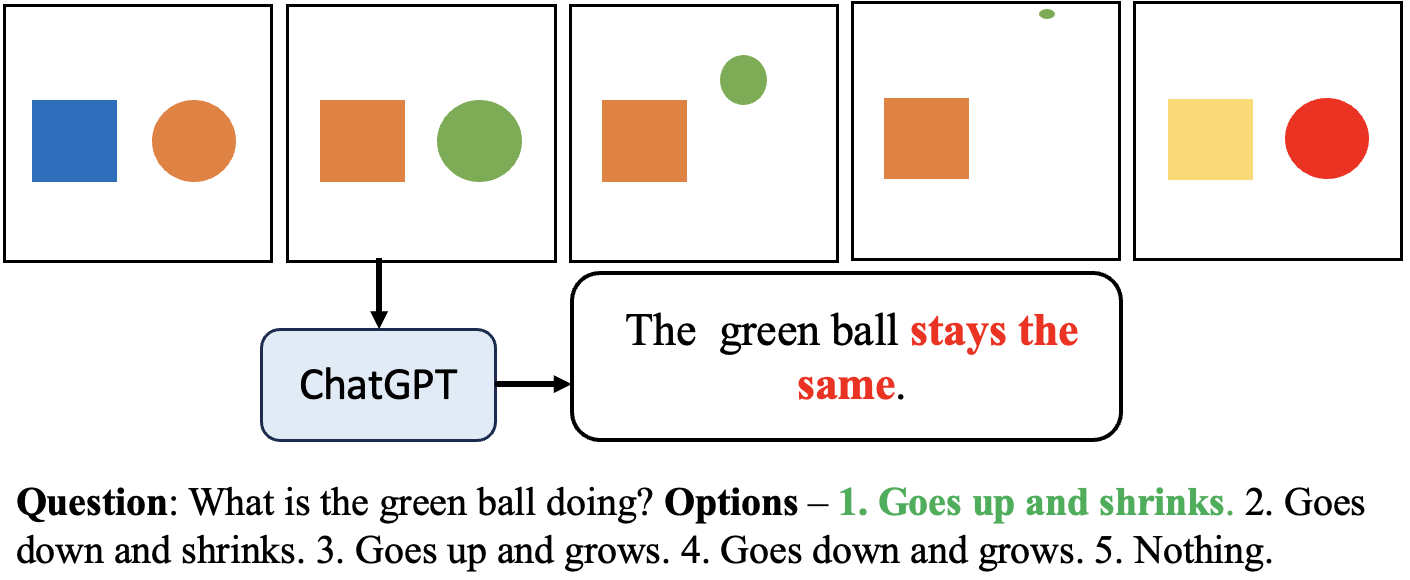}
    \subcaption{Dynamic event: target manipulation.}
  \end{minipage}
  \hfill

  \caption{Different types of events of \oursben.}
 \label{fig: video_pic}
\end{figure*}

\oursben  is  a new benchmark for studies of how complexity attributes affect video-LLM performance. It is inspired by the well established protocols for the study of visual attention~\cite{craik2014effects, geng2014attentional, lavie2010attention, smith2024distractors, wang2025dual, lorenc2021distraction}, relying on a video dataset of simple but controllable objects, namely colored balls and squares displayed against a white background, as shown in Figure~\ref{fig: video_pic}. Both balls and squares are subject to attribute changes, \eg a change of size or color. 

 The main benefit of \oursben over benchmarks of real video is that each attribute can be manipulated {\it independently\/}. This is important to show that the attribute can, {\it by itself\/}, determine video complexity. In every experiment of the current implementation of \oursben, the target object is a green ball. The video-LLM is asked a question $q$ grounded on an event ${\cal E}(q)$, which is an {\it attribute change\/} of the target, \eg color in Figure~\ref{fig: video_pic} (a), and presented with 5 possible answers. All remaining squares in the video serve as distractors that increase the difficulty of detecting the event. Two types of events are defined and a pair of target and distractor manipulations defined for each.

\noindent{\bf Static events.} In these events, the target manipulation involves switching the value of a static attribute, namely the color, of the target. As shown in Figure~\ref{fig: video_pic} a) the event $\mathcal{E}$ is a change of the color of the green ball at a unique {\it event location} (frame pair) in the video. The remaining balls and squares are chosen randomly (restricting ball colors to non-green) per frame pair. The target manipulation consists of varying the {\it event location} in the video. This can be  a) beginning of video, $j \in [0,2 \gamma]$, b) middle of video, $j \in [4 \gamma, 6 \gamma]$ and c) end of video, $j \in [8 \gamma, 10 \gamma]$. The task is defined by a question that asks the video-LLM to choose the new color of the ball from $5$ options. In the example of Figure~\ref{fig: video_pic} a) the correct answer is `red'. The distractor manipulation consists of varying the {\it non-event length}, by concatenating frames at the end of the video, so that $N \in [10 \gamma, 120 \gamma]$. As the video length increases, the event becomes more of a needle in haystack, and the task is harder.

\noindent{\bf Dynamic events.} In this case, the target manipulations are dynamic, namely the combination of (upward or downward) ball motion and variation (increase or decrease) of ball size. Figure~\ref{fig: video_pic} c) shows an example where the ball moves up and shrinks. The task is to choose from one of the four possible combinations of motion and shape change plus the `none of the above' option. The distractor manipulation consists of increasing the {\it number of distractors}. This allows the appearance of the entire scene to change across frame pairs, creating a dynamic background and increasing the difficulty of finding the target, especially as the number of distractors grows, as illustrated in Figure~\ref{fig: video_pic} b). The target manipulation is to increase the {\it rate of change of the target\/}. This is implemented by varying the {\it frame rate\/} $\gamma$.  Given an event frame rate $\gamma_E$, the Nyquist rate $\gamma_{N} = 2 \times \gamma_{E}$ is a lower bound for the video frame rate, in the sense that for any $\gamma < \gamma_{N}$ it should be hard to understand the video even for a human. We used $\gamma_E = 2$ and varied $\gamma$ between 1 and 32. The event position is always at the beginning of video ($j \in [0, 2 \gamma]$) in these experiments. 

Overall, the \oursben currently allows {\it independent} control of four {\it attributes of video complexity}: event location ${\cal A}_1$, surrounding (non-event) video length ${\cal A}_2$, scene complexity (number of distractors) ${\cal A}_3$, and event speed (frame rate) ${\cal A}_4$. These are proxies for real video event attributes such as the length of an informative action versus the length of the video, action speed, or the complexity of the background scene. While these properties are difficult to control in real videos, the hypothesis of the paper is that these attributes affect \emph{temporal complexity} and thus video-LLM performance.
\oursben allows detailed studies of this hypothesis. Note that we make no claims that these are the only complexity attributes of video. In fact, we propose \oursben as an evolving benchmark, to which other researchers interested in studies of video complexity can add different attribute hypothesis. 

\noindent{\bf Dataset.} \oursben comprises 1200 video-question pairs, with 700 featuring static events and 500 containing dynamic events, generated as follows. For static events, 300 videos include target manipulations, where the event is equally distributed between  beginning, middle and end of the video. For distractor manipulation, 400 videos were generated with length $N \in \{20, 40, 60, 80\}$, with 100 videos per video length. For dynamic events, target manipulations are based on 100 videos with frame rate of $2$ fps. The manipulations consisted of adjusting the fps setting  of the video-LLM to a value $\gamma \in \{0.8, 1.6, 3.2, 4.8, 6.4, 8\}$. For distractor manipulations, 400 videos were created with $D\in \{1, 2, 3, 4\}$ distractors, with 100 videos per category.

\end{document}